\documentclass[10pt,twocolumn,letterpaper]{article}

\usepackage[pagenumbers]{cvpr} 
\usepackage{array}
\usepackage{times}
\usepackage{epsfig}
\usepackage{graphicx}
\usepackage{amsmath}
\usepackage{amssymb}
\usepackage{verbatim}
\usepackage{hhline}
\usepackage{multirow}
\usepackage{makecell}

\usepackage{xcolor,colortbl}

\usepackage{xspace}

\usepackage{amssymb}
\usepackage{pifont}
\usepackage{subcaption}

\makeatletter
\DeclareRobustCommand\onedot{\futurelet\@let@token\@onedot}
\def\@onedot{\ifx\@let@token.\else.\null\fi\xspace}
\def\eg{\emph{e.g}\onedot} 
\def\ie{\emph{i.e}\onedot}

\newcommand{\Tref}[1]{Table~\textcolor{blue}{\ref{#1}}}

\newcommand{\Fref}[1]{Fig.~\textcolor{blue}{\ref{#1}}}

\newcommand{\Aref}[1]{Alg.~\textcolor{blue}{\ref{#1}}}

\DeclareMathOperator*{\argmax}{arg\,max}

\usepackage{graphicx}
\usepackage{amsmath}
\usepackage{amssymb}
\usepackage{booktabs}
\usepackage{soul}
\usepackage{bbm}
\usepackage{enumitem}

\makeatletter
\@namedef{ver@everyshi.sty}{}
\makeatother
\usepackage{tikz}

\usepackage{tikz}
\usetikzlibrary{arrows,intersections}
\tikzset{font=\scriptsize}
\usepackage{pgfplots}
\pgfplotsset{compat=1.11}
\usepgfplotslibrary{fillbetween}
\usetikzlibrary{backgrounds}

\definecolor{plotsgreen}{RGB}{38,150,38}
\definecolor{plotsorange}{RGB}{255,115,17}
\definecolor{plotsred}{RGB}{208,34,35}
\definecolor{plotsyellow}{RGB}{255,225,80}
\definecolor{plotsgrey}{RGB}{128,128,128}
\definecolor{plotspurple}{RGB}{137,92,181}
\definecolor{plotsblue}{RGB}{0,101,167}

\usepackage[pagebackref,breaklinks,colorlinks,linkcolor=blue,citecolor=blue]{hyperref}

\usepackage[capitalize]{cleveref}
\crefname{section}{Sec.}{Secs.}
\Crefname{section}{Section}{Sections}
\Crefname{table}{Table}{Tables}
\crefname{table}{Tab.}{Tabs.}
\usepackage[linesnumbered,ruled,vlined]{algorithm2e}
\SetKwInOut{Parameter}{Parameters}
\usepackage{nicematrix}

\def\cvprPaperID{4586} 
\def\confName{CVPR}
\def\confYear{2022}

\begin{document}

\title{MM-TTA: Multi-Modal Test-Time Adaptation for 3D Semantic Segmentation}

\author{Inkyu Shin$^{1 }$ \quad Yi-Hsuan Tsai$^{2 }$ \quad Bingbing Zhuang$^{3 }$ \quad Samuel Schulter$^{3 }$ \\
Buyu Liu$^{3 }$ \quad Sparsh Garg$^{3 }$ \quad In So Kweon$^{1 }$ \quad Kuk-Jin Yoon$^{1 }$\\
$^{1}$KAIST \quad \ $^{2}$Phiar \quad \ $^{3}$NEC Laboratories America
}


\maketitle

\begin{abstract}
Test-time adaptation approaches have recently emerged as a practical solution for handling domain shift without access to the source domain data. In this paper, we propose and explore a new multi-modal extension of test-time adaptation for 3D semantic segmentation.
%
We find that, directly applying existing methods usually results in performance instability at test time, because multi-modal input is not considered jointly. To design a framework that can take full advantage of multi-modality, where each modality provides regularized self-supervisory signals to other modalities, we propose two complementary modules within and across the modalities.
First, Intra-modal Pseudo-label Generation (\textbf{Intra-PG}) is introduced to obtain reliable pseudo labels within each modality by aggregating information from two models that are both pre-trained on source data but updated with target data at different paces.
Second, Inter-modal Pseudo-label Refinement (\textbf{Inter-PR}) adaptively selects more reliable pseudo labels from different modalities based on a proposed consistency scheme. 
Experiments demonstrate that our regularized pseudo labels produce stable self-learning signals in numerous multi-modal test-time adaptation scenarios for 3D semantic segmentation. 
Visit our project website at \url{https://www.nec-labs.com/~mas/MM-TTA}

\end{abstract}

\section{Introduction}



3D semantic segmentation is a challenging task that requires both geometric and semantic reasoning about the input scene, but it can provide rich insights that enable applications like autonomous driving~\cite{wu2017squeezeseg,xu2021rpvnet}, virtual reality and robotics~\cite{choy20194d,thomas2019kpconv}.
With the advancement of sensor technology, multi-modal sensors are considered as the key to effectively tackle this task~\cite{meyer2019sensor,madawy2019rgb,iterative}.  In particular, to obtain more accurate 3D point-level semantic understanding, contextual information in 2D RGB images can be reinforced by the geometric property of 3D points from LiDAR sensors, and vice versa.  Therefore, it is of great interest to develop multi-modal approaches for 3D semantic segmentation.

\begin{figure}
\begin{center}
\includegraphics[width=0.75\linewidth]{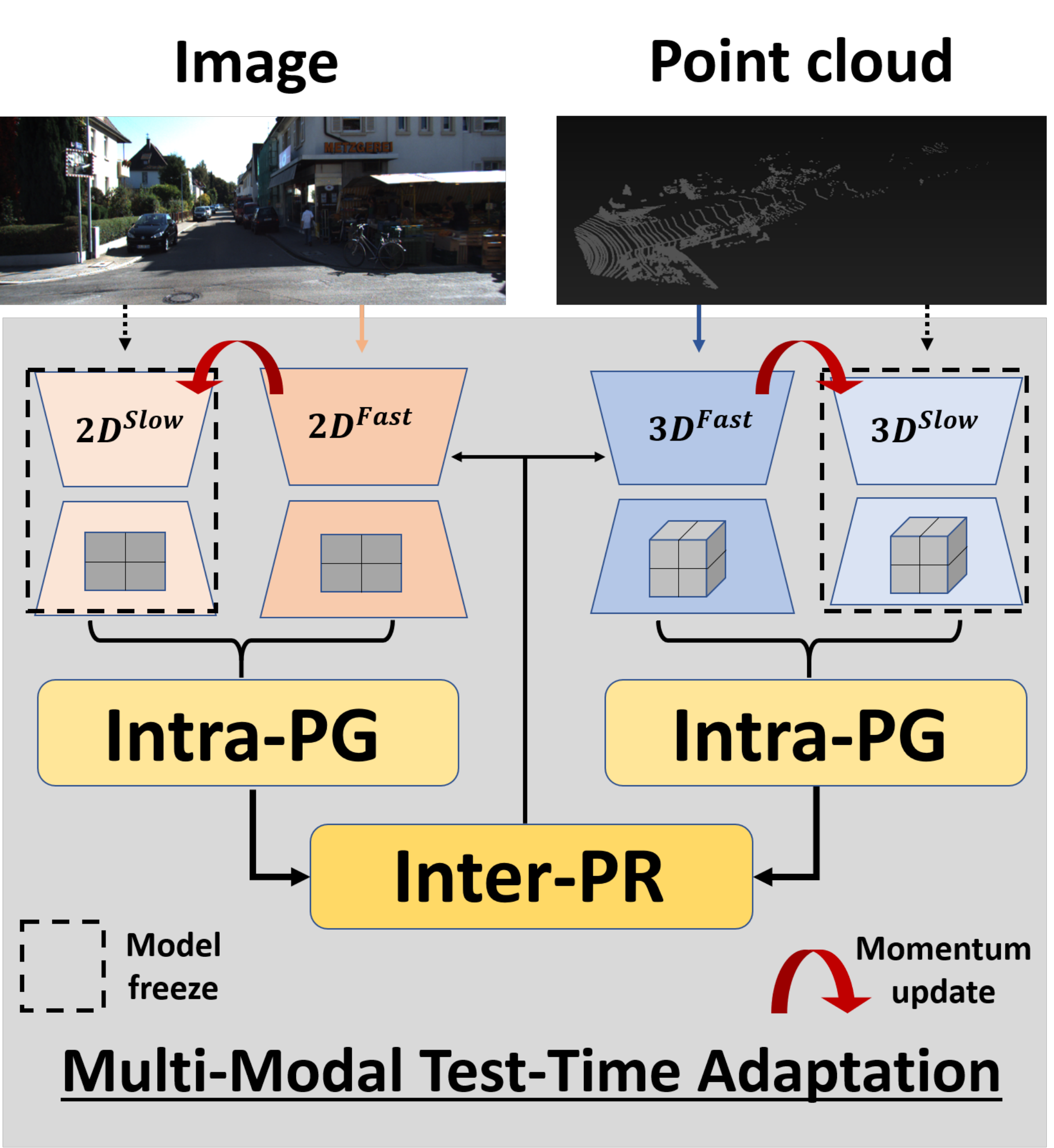}
\caption{
We propose a Multi-Modal Test-Time Adaptation (MM-TTA) framework that enables a model to be quickly adapted to multi-modal test data without access to the source domain training data.
We introduce two modules: 1) Intra-PG to produce reliable pseudo labels within each modality via updating two models (batch norm statistics) in different paces, \ie, slow and fast updating schemes with a momentum, and 2) Inter-PR to adaptively select pseudo-labels from the two modalities. These two modules seamlessly collaborate with each other and co-produce final cross-modal pseudo labels to help test-time adaptation.
}
\label{fig:teaser}
\end{center}
\vspace{-5mm}
\end{figure}

However, multi-modal data is sensitive to a distribution shift at test time when a domain gap exists to the training data~\cite{Bayoudh2021ASO}.
Therefore, it is critical for a model to quickly adapt to the new multi-modal data during testing for obtaining better performance, \ie, through test-time adaptation (TTA)~\cite{wang2021tent,prabhu2021s4t}.
This is different from the usual domain adaptive semantic segmentation setting~\cite{tsai2020learning,yi2021complete,jaritz2020xmuda} that can access both source and target data during training.
In TTA, we only have access to model parameters pre-trained on the source data and the unlabeled test data for \emph{quick} adaptation, which typically (and also in this work) refers to one epoch of training.
This is practical for real-world scenarios, but it is also challenging because only the target data is available with a limited budget for adaptation.



In this paper, we study multi-modal 3D semantic segmentation in the setting of test-time adaptation, using both image and point cloud as input.
Prior works on general test-time adaptation like TENT~\cite{wang2021tent} propose entropy minimization as a self-training loss to update batch norm parameters.
While TENT~\cite{wang2021tent} is not designed for multi-modality, we show a simple extension that updates parameters in individual branches for each modality (2D image and 3D point cloud).
%
However, we find that this extension causes instability during training. One reason is that, since entropy minimization tends to generate sharp output distributions, using it separately for 2D and 3D branches may increase the cross-modal discrepancy. This would further lead to a sub-optimal model ensemble for 2D and 3D outputs, which is the common scheme for multi-modal semantic segmentation.
One way to alleviate this cross-modal discrepancy is to utilize a consistency loss~\cite{jaritz2020xmuda} between predictions of 2D and 3D branches, via KL divergence. However, since the test data during adaptation is unlabeled, enforcing the consistency across modalities may even worsen predictions if the output of one branch is inaccurate.

To tackle the aforementioned issues and design better test-time self-supervisory signals, we propose a
cross-modal regularized self-training framework that aims to generate reliable and adaptive pseudo labels (see \Fref{fig:teaser}).
Our method mainly consists of two modules: 1) Intra-modal Pseudo-label Generation (\textbf{Intra-PG}), and 2) Inter-modal Pseudo-label Refinement (\textbf{Inter-PR}).
For the intra-modal module, we aim to produce reliable pseudo labels in each modality that alleviate the instability issue in test-time adaptation, \ie, only updating batch norm parameters by seeing the test data once. To this end, we design a slow-fast modeling strategy.
Specifically, to maintain the model stability, we initialize one batch norm statistics from the pre-trained source model, and \textit{slowly} update it with a momentum from another fast-updated batch norm parameter, while this \textit{fast}-updated model is directly updated by the test data, which is more aggressive but also provides up-to-date statistics.
Our model is thus able to fuse predictions from the slow-/fast-updated statistics to enjoy their complementary benefits. 

For the inter-modal module, we propose to adaptively select reliable pseudo labels from the individual 2D and 3D branches, because each modality brings its own advantage for 3D semantic segmentation.
To this end, we first leverage the Intra-PG module to measure the prediction consistencies of each modality separately, and then provide a fused prediction from slow-fast models to the Inter-PR module (\Fref{fig:teaser}).
Based on these consistencies, our model adaptively selects reliable pseudo labels from two modalities to form a final cross-modal pseudo label as the self-training signal to update 2D/3D batch norm parameters.

The proposed two modules collaborate with each other for multi-modal test-time adaptation, and thus we name our framework as \textit{MM-TTA}.
We conduct extensive experiments to include several TTA state-of-the-art baselines and show that our MM-TTA framework achieves favorable performance over different benchmark settings, including cross-dataset with different sensors, synthetic-to-real, and day-to-night scenarios.
%
Moreover, we provide comprehensive analysis to demonstrate the benefits of our two proposed modules (Intra-PG and Inter-PR) and the stability comparisons with existing methods.
Here are our main contributions:
\begin{enumerate}
\vspace{-1mm}
\setlength\itemsep{0.3em}
    \item We explore a new task, test-time adaptation for multi-modal 3D semantic segmentation, and propose a framework that effectively produces cross-modal
    pseudo labels as self-training signals.

    
    \item We introduce two modules that seamlessly work together: The Intra-PG module produces pseudo labels for each modality separately and the Inter-PR module adaptively selects pseudo labels across modalities.
    
    
    \item We demonstrate our framework under different adaptation settings with extensive ablation studies and experimental comparisons against strong baselines and state-of-the-art methods.
\end{enumerate}




\section{Related Work}


\noindent\textbf{Test-Time Adaptation (TTA)} aims to enable quick adaptation of an existing model to new target data without having access to the source domain data the model was trained on.
As an important challenge for dealing with dynamic domain shift in real-world, TTA is attracting more and more attention in several tasks~\cite{sun2020testtime, wang2021tent, prabhu2021s4t, Chi2021TestTimeFA, li2021testtime}. Among them, Test-time Training (TTT)~\cite{sun2020testtime} updates model parameters in an online manner by applying a self-supervised proxy task on the test data. Since this proxy task is also required for training samples, finding an optimal proxy task that works well in both training and testing is challenging.

From that point of view, TENT~\cite{wang2021tent}, the first Test-Time Adaptation (TTA) approach, proposes a simple yet effective entropy minimization method to optimize for test-time batch norm parameters without requiring any proxy task during training, which is demonstrated for image classification and 2D semantic segmentation.
However, entropy minimization tends to encourage the model to increase confidence despite false predictions. To design a regularized self-learning signal at test-time, a concurrent work, S4T~\cite{prabhu2021s4t}, proposes a selective self-training scheme for 2D semantic segmentation by regularizing pseudo labels with aligned predictive view generation. Nonetheless, this design is considered to be specific to an image-level task where spatial augmentation can be performed. 
Compared to the aforementioned work, we study a similar TTA setting but in the different context of using multi-modality for 3D semantic segmentation, \ie, Multi-Modal Test-Time Adaptation (MM-TTA), in which we develop intra-modal and inter-modal modules that seamlessly work with each other to obtain more reliable self-learning signals.

\vspace{2mm}
\noindent\textbf{3D Semantic Segmentation} has been recognized as an important 3D scene understanding task aimed at classifying each LiDAR point into semantic categories.
Therefore, point clouds from LiDAR are deemed to be the dominant modality to solve this task~\cite{wu2017squeezeseg, wang2018pointseg, thomas2019kpconv, choy20194d, xu2021rpvnet}.
Range-based methods~\cite{wu2017squeezeseg, wang2018pointseg} adopt spherical projection to project 3D points onto the 2D image plane and then pass this through a 2D-based backbone~\cite{iandola2016squeezenet}.
%
Another effort is to utilize the raw 3D point cloud by designing 3D segmentation models~\cite{thomas2019kpconv,choy20194d}. KPConv~\cite{thomas2019kpconv} operates on point clouds without any intermediate representation, while MinkowskiNet~\cite{choy20194d} voxelizes the point cloud and utilizes SparseConvNet~\cite{SubmanifoldSparseConvNet} for processing. Despite these efforts, the LiDAR point cloud itself lacks 2D contextual information that is essential to understand the complex semantics of a scene.

To address this weakness,
recent work~\cite{meyer2019sensor, iterative} explores the use of multi-modal inputs (RGB images and LiDAR point clouds) for 3D segmentation. These methods commonly separate the backbones for 2D and 3D modalities and propose fusion techniques between the two outputs. Considering both contextual and geometric information from each modality is shown to boost the performance in 3D semantic segmentation. However, since each modality has different dataset biases (\eg, style distribution in 2D and point distribution in 3D), multi-modality based models are harder to adapt to new data.
In this work, different from supervised training, we tackle multi-modal 3D semantic segmentation in the test-time adaptation setting, which is practical as it incorporates test data statistics during inference, thus improving results from multi-modal baselines. 



\vspace{2mm}
\noindent\textbf{Unsupervised Domain Adaptation (UDA)}
aims at bridging the gap between labeled source data and unlabeled target data.
Methods for both 2D~\cite{tsai2020learning, vu2018advent, zou2018domain, zou2020confidence, shin2020two,kim2021learning} and 3D~\cite{cross_sensor, saleh2019domain, wu2018squeezesegv2, yi2021complete} data have been proposed.
Recently, few works \cite{jaritz2020xmuda, peng2021sparsetodense} also introduced UDA approaches for 2D/3D multi-modal data.
Specifically, xMUDA~\cite{jaritz2020xmuda} executes consistency learning at training time between two modalities both in source and target domains, while DsCML~\cite{peng2021sparsetodense} further utilizes adversarial learning with dynamic sparse-to-dense cross-modal learning between modalities. 
All UDA methods are allowed to access source data during adaptation, while we tackle test-time adaptation, in which only the source pre-trained model is available and a limited budget is given to update test time statistics of the model.
%



\begin{figure*}
\begin{center}
\includegraphics[width=0.94\linewidth]{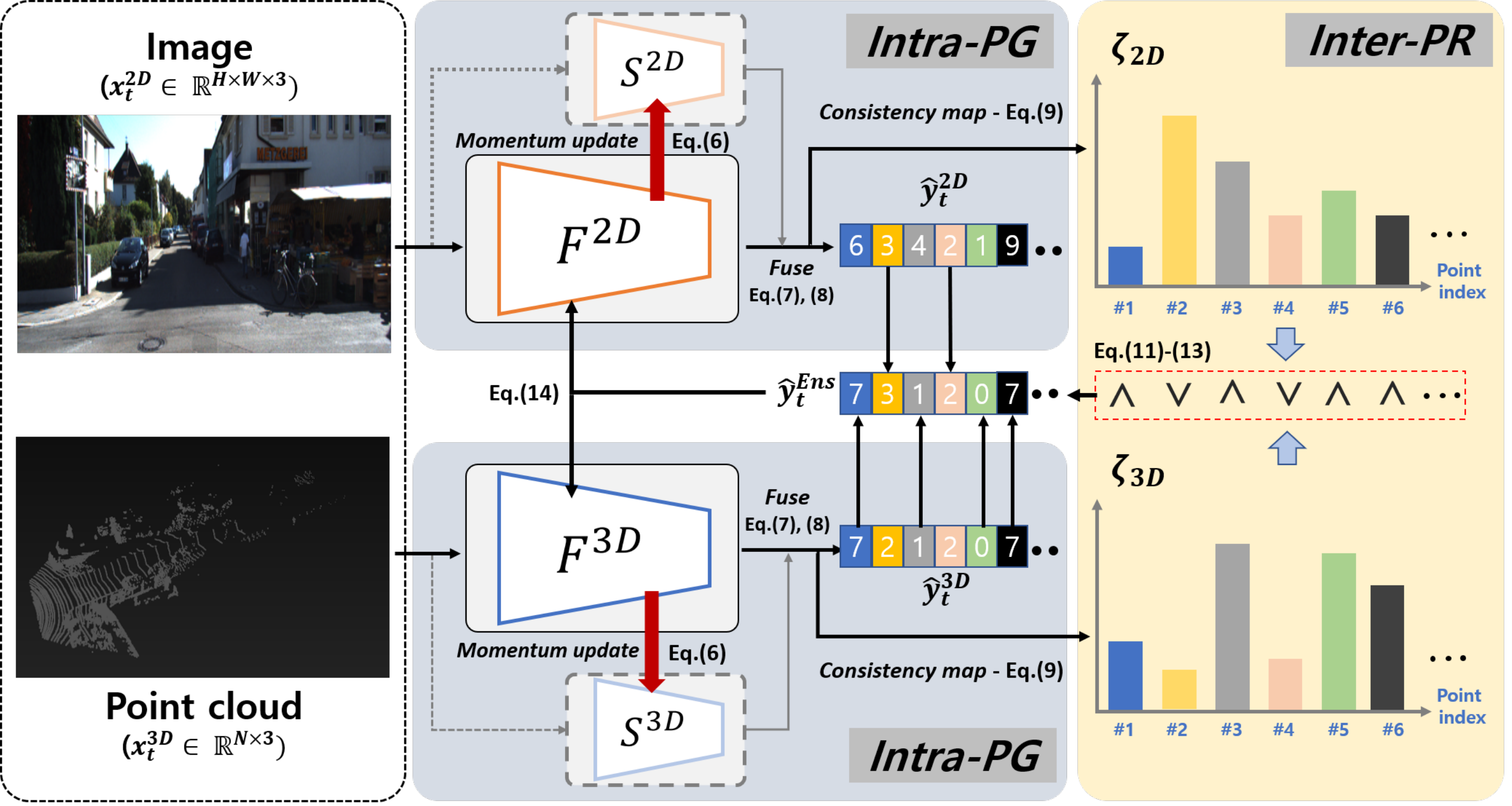}
\caption{\textbf{Overview of the proposed Multi-Modal Test-Time Adaptation (MM-TTA) framework.} Our MM-TTA consists of two modules: Intra-modal Pseudo-label Generation (\textbf{intra-PG}) and Inter-modal Pseudo-label Refinement (\textbf{inter-PR}). For Intra-PG, we adopt a slowly-updated model $S$ that is gradually updated by a fast-updated model $S$ with a momentum. Note that, statistics in the fast-updated model $S$ are directly updated by the data, which is more aggressive but up-to-date, while the model $S$ slowly moves towards the target data statistics and thus is more stable.
By aggregating slow-fast models, each modality can generate robust pseudo labels ($\hat{y}^{2D}_{t}$ and $\hat{y}^{3D}_{t}$). For Inter-PR, we measure the consistency map between slow-fast models and enable an adaptive selection process for finding confident pseudo labels based on calculated $\zeta_{2D}$ and $\zeta_{3D}$. After obtaining the cross-modal regularized pseudo label ($\hat{y}^{Ens}_{t}$) by jointly considering 2D and 3D confidences, we update the batch norm parameters for $F$ in both modalities.
}
\label{fig:network}
\end{center}
\vspace{-5mm}
\end{figure*}

\section{Proposed Method}
We start by introducing the preliminaries for test-time adaptation in~\cref{sec:preliminary}.
Then, we explore several baselines for multi-modal test-time adaptation in~\cref{sec:baseline}, using the image (2D modality) and point cloud (3D modality) as input for 3D semantic segmentation.
Finally, we propose our Multi-Modal Test-Time-Adaptation (MM-TTA) framework, as shown in~\cref{fig:network}, with two newly designed modules: 1) Intra-PG to generate pseudo-labels within each modality (\cref{sec:intra_PG}), and 2) Inter-PR to adaptively select reliable pseudo-labels across modalities (\cref{sec:inter_PR}).


\vspace{2mm}
\noindent\textbf{Setup and notation.}
We follow the setting of test-time adaptation \cite{wang2021tent}, where we are not able to access the source data but only the source pre-trained multi-modal segmentation model. This model consists of 2D and 3D branches, $F^{\textrm{2D}}$ and $F^{\textrm{3D}}$, each of which including the feature extractor $G^{\textrm{2D}}/G^{\textrm{3D}}$ and a classifier.
%
Here, we also denote the multi-modal test-time target data (see \Fref{fig:network}), images $x_{t}^{\textrm{2D}}\in \mathbb{R}^{H \times W \times 3}$ and point clouds $x_{t}^{\textrm{3D}} \in \mathbb{R}^{N \times 3}$ (3D points in the camera field of view).
Note that the feature extracted from the 2D branch, $G^{\textrm{2D}}(x_{t}^{\textrm{2D}}) \in \mathbb{R}^{H \times W \times f}$, is sampled
at the $N$ projected 3D points resulting in a feature shape of $N \times f$.
Individual network predictions from 2D/3D are denoted as $p(x_t^M) = F^M(x_t^M) \in \mathbb{R}^{N \times |K|}$, where $|K|$ is the number of categories and $M \in \{\textrm{2D},\textrm{3D}\}$.
%




\subsection{Preliminaries}
\label{sec:preliminary}
\noindent\textbf{Batch Normalization (BN)}~\cite{pmlr-v37-ioffe15} has been widely used in current DNNs for both 2D and 3D models. It generally includes normalization statistics and transformation parameters in the $j$-th BN layer given the target mini-batch input
$x_{t}^{M}$ with $M \in \{\textrm{2D},\textrm{3D}\}$:
\begin{equation}%
\begin{split}
\hat{x}^{M}_{t_{j}} = \frac{x^{M}_{t_{j}} - \mu^{M}_{t_{j}}}{\sigma^{M}_{t_{j}}} \quad \textrm{and} \quad y^{M}_{t_{j}} = \gamma^{M}_{t_{j}}\hat{x}^{M}_{t_{j}}+ \beta^{M}_{t_{j}},
\end{split}
\label{eqn:BN}
\end{equation}
where $\mu^{M}_{t_{j}} = \mathop{{}\mathbb{E}}[x^{M}_{t_{j}}]$ and $(\sigma^{M}_{t_{j}})^{2} = \mathop{{}\mathbb{E}}[(\mu^{M}_{t_{j}} - x^{M}_{t_{j}})^{2}]$ are normalization statistics, and $\gamma^{M}_{t_{j}}, \beta^{M}_{t_{j}}$ are learnable transformation parameters.
To simplify notation, we use $\Omega^{\textrm{2D}}_{t} = (\mu, \sigma, \gamma, \beta)^{\textrm{2D}}_{t}$ for 2D and $\Omega^{\textrm{3D}}_{t} = (\mu, \sigma, \gamma, \beta)^{\textrm{3D}}_{t}$ for 3D.

\vspace{2mm}
\noindent\textbf{The number of parameters updated in Test-time Adaptation} is constrained to be small for reasons of efficiency and stability.
Following TENT~\cite{wang2021tent}, we only estimate and optimize $(\mu, \sigma, \gamma, \beta)_{t}$ occupying $<$1\% of parameters both in 2D and 3D branches. 

\subsection{Baselines for MM-TTA}
\label{sec:baseline}
Since we propose the first attempt of multi-modal test-time adaptation for 3D semantic segmentation, we first study several self-learning baselines based on existing methods that we extend to our MM-TTA setting.

\vspace{2mm}
\noindent\textbf{Self-learning with Entropy} is originally proposed by TENT~\cite{wang2021tent}. Its test-time objective $L(x_{t})$ is to minimize the entropy of model predictions $p(x^{M}_{t}) = F^{M}(x^{M}_{t})$, where $F^M$ is either the 2D or 3D branch (recall that $M \in \{\textrm{2D},\textrm{3D}\}$).
The overall objective of entropy minimization for this MM-TTA baseline is expressed as:
\begin{equation}
\begin{split}
    L_{\textrm{ent}}(x_{t}) = - \sum_{k}p(x^{\textrm{2D}}_{t})^{(k)}\log p(x^{\textrm{2D}}_{t})^{(k)} \\
    - \sum_{k}p(x^{\textrm{3D}}_{t})^{(k)}\log p(x^{\textrm{3D}}_{t})^{(k)},
\end{split}
\label{eqn:ent}
\end{equation}
where $k$ denotes the class.
Despite its simplicity, this objective only encourages sharp output distributions, which may reinforce wrong predictions, and may not lead to cross-modal consistency.

\vspace{2mm}
\noindent\textbf{Self-learning with Consistency} aims to achieve multi-modal test-time adaptation via a consistency loss between predictions of 2D and 3D modalities:
\begin{equation}
\begin{split}
    L_{\textrm{cons}}(x_{t}) = D_{\textrm{KL}}(p(x^{\textrm{2D}}_{t})||p(x^{\textrm{3D}}_{t})) \; \\
    + \; D_{\textrm{KL}}(p(x^{\textrm{3D}}_{t})||p(x^{\textrm{2D}}_{t})),
\end{split}
\label{eqn:cons}
\end{equation}
where $D_{\textrm{KL}}$ is the KL divergence. Different from xMUDA~\cite{jaritz2020xmuda}, which operates in the standard domain adaptation setting with access to the source data, our MM-TTA is not regularized by the source task loss and thereby this objective may fail to capture the correct consistency when one of the branches provides a wrong prediction.

\vspace{2mm}
\noindent\textbf{Self-learning with Pseudo-labels} is another common approach for test-time adaptation. Typically, pseudo-labels $\hat{y}_{t}$ can be obtained by:
\begin{equation}
  \hat{y}_{t} = \argmax_{k \in K} \mathbbm{1}[ \; p(x_t)^{(k)} > \theta^{(k)} \; ] \; p(x_t)^{(k)},
  \label{eq:cut}
\end{equation}
where $\mathbbm{1}[\cdot]$ is an indicator function returning true if the condition is satisfied, \ie, if prediction $p(x_t)^{(k)}$ for class $k$ is larger than the threshold $\theta^{(k)}$. Note that the pseudo-label $\hat{y}_{t}$ can be obtained similarly for both 2D and 3D branches, \ie, $\hat{y}^{\textrm{2D}}_{t}$ and $\hat{y}^{\textrm{3D}}_{t}$. The objective for pseudo-labeling uses the standard cross-entropy loss $L_{\textrm{seg}}$ for semantic segmentation:
\begin{equation}
\begin{split}
    L_{\textrm{pseudo}}(x_{t}) = L_{\textrm{seg}}(p(x^{\textrm{2D}}_{t}),\hat{y}^{\textrm{2D}}_{t}) \; \\
    + \; L_{\textrm{seg}}(p(x^{\textrm{3D}}_{t}),\hat{y}^{\textrm{3D}}_{t}).
\end{split}
\label{eqn:pseudo}
\end{equation}
Although pseudo-labels provide supervisory signals to update models, there are potential issues when it is applied to our MM-TTA setting.
First, only the batch norm statistics are updated to replace the original source statistics during adaptation, but the model to generate pseudo labels for target data still mainly consists of fixed parameters pre-trained on source data, which can lead to low-quality pseudo labels.
Second, the model still lacks information exchange across modalities to refine pseudo labels, which can also result in sub-optimal performance.
In contrast, our proposed MM-TTA framework provides simple yet effective solutions to these limitations with the following two modules.


\subsection{Intra-modal Pseudo-label Generation}
\label{sec:intra_PG}
We propose Intra-PG to generate reliable online pseudo labels within each modality by having two models, $S^M$ and $F^M$, with different updating paces (see \Fref{fig:network}).
%
First, we define a fast-updated model $F^M$ that replaces and updates batch norm statistics directly from the test data, which is identical to baseline models in Section~\ref{sec:baseline}.
%
Second, we introduce an additional slowly-updated model $S^M$ that is initially source pre-trained and has a momentum update scheme from the fast-updated model $F^M$. In short, we denote these two models as slow/fast model as $S^M$/$F^M$.
%
That is, the statistics in the fast model are updated more aggressively by the test data, while the slow model's statistics gradually move towards the target statistics, and thus provide a stable and complementary supervisory signal.
Note that only the slow model $S^M$ is used at inference time.
Here, we present the batch norm statistics for the slow model $S^M$ as:
%
\begin{equation}
\begin{split}
  \Omega^{S}_{t_{i}} = (1-\lambda)\Omega^{F}_{t_{i}} + \lambda\Omega^{S}_{t_{i-1}}, \\
  \Omega^{S}_{t_{0}} = \Omega_{s},
  \label{eq:intra_1}
\end{split}
\end{equation}
%
%
where $\Omega^{S}_{t_{i}} = (\mu, \sigma, \gamma, \beta)^{S}_{t_{i}}$ is the moving averaged statistics at iteration $i$ with a momentum factor $\lambda$ to aggregate fast model's statistics $\Omega^{F}_{t_{i}}$ and slow model's statistics $\Omega^{S}_{t_{i-1}}$. The initial statistics $\Omega^{S}_{t_{0}}$ are from the source pre-trained model denoted as $\Omega_{s}$.
Note that, when we set a large value for $\lambda$ (0.99 in the paper), it will move slower towards the target statistics, and otherwise it moves faster.
To further leverage both the slow-fast statistics in each modality, we fuse their predictions as:
%
\begin{equation}
  p(x^{M}_{t}) = \frac{(S^{M}(x^{M}_{t}) + F^{M}(x^{M}_{t}))}{2}.
  \label{eq:intra_2}
\end{equation}
Then, we can obtain aggregated pseudo labels from slow-fast models for each modality $M \in \{\textrm{2D},\textrm{3D}\}$:
\begin{equation}
  \hat{y}^{M}_{t}= \argmax_{k \in K} p(x^{M}_{t})^{(k)}.
  \label{eq:intra}
\end{equation}

\subsection{Inter-modal Pseudo-label Refinement}
\label{sec:inter_PR}

After obtaining initial aggregated pseudo labels for each modality in \eqref{eq:intra}, we propose the Inter-PR module to improve pseudo labels via cross-modal fusion.
To realize this, we first calculate a consistency measure ($\zeta_{M}$) between slow and fast models of Intra-PG for each modality separately: 
\begin{equation}
\zeta_{\textrm{M}} = Sim(S^{M}(x^{M}_{t}), F^{M}(x^{M}_{t})),
\label{eq:cons}
\end{equation}
where we define $Sim(\cdot)$ as the inverse of KL divergences to express the similarity between two probabilities:
\begin{equation}
  Sim(x,y) = \left(\frac{1}{D_{\textrm{KL}}(x||y)+\epsilon} + \frac{1}{D_{\textrm{KL}}(y||x)+\epsilon}\right) / 2.
  \label{eq:sim}
\end{equation}
Here, $\epsilon$ is a small scalar constant to prevent division-by-zero.
This consistency measure helps us to fuse the per-modality predictions and estimate more reliable pseudo labels.
%
We propose two variants: \textit{Hard Select} and \textit{Soft Select}.
The former takes each pseudo label exclusively from one of the modalities, while the latter conducts a weighted sum of pseudo labels from the two modalities using the consistency measure.
%
We define \textit{Hard Select} as
\begin{equation}
\begin{split}
    \hat{y}_{t}^{H} = \begin{cases}
    \hat{y}^{\textrm{2D}}_{t}, & \text{if} \zeta_{\textrm{2D}} \geq \zeta_{\textrm{3D}}, \\
    \hat{y}^{\textrm{3D}}_{t}, & \text{otherwise}. \\ 
    \end{cases}
    \label{eq:hard}
\end{split}
\end{equation}
and \textit{Soft Select} as
\begin{equation}
    \hat{y}_{t}^{S} = \argmax_{k \in K}p^{W(k)}_{t},
    \label{eq:soft_1}
\end{equation}
%
%
with $p^{W(k)}_{t} = \zeta_{\textrm{2D}}^{\ast} \: p(x^{\textrm{2D}}_{t})^{(k)} + \zeta_{\textrm{3D}}^{\ast} \: p(x^{\textrm{3D}}_{t})^{(k)}$ and
%
where $\zeta_{\textrm{2D}}^{\ast} = \zeta_{\textrm{2D}} / (\zeta_{\textrm{2D}}+\zeta_{\textrm{3D}})$, and $\zeta_{\textrm{3D}}^{\ast} = 1 - \zeta_{\textrm{2D}}^{\ast}$ are normalized consistency measures.
In addition, we ignore pseudo labels whose maximum consistency measure over the two modalities, \ie, $\max(\zeta_{\textrm{2D}},\zeta_{\textrm{3D}})$, is below a threshold $\theta^{(k)}$.
%
Formally, our MM-TTA objective to use the generated pseudo label $\hat{y}_{t}^{\textrm{Ens}}$ ($\hat{y}_{t}^{H}$ or $\hat{y}_{t}^{S}$) for updating batch norm statistics is:
\begin{equation}
L_{\textrm{mm-tta}}(x_{t}) = L_{\textrm{seg}}(p(x^{\textrm{2D}}_{t}),\hat{y}_{t}^{\textrm{Ens}}) + L_{\textrm{seg}}(p(x^{\textrm{3D}}_{t}),\hat{y}_{t}^{\textrm{Ens}}).
\label{eqn:mm-tta}
\end{equation}
%


\section{Experimental Results}
\begin{table*}
\centering
\resizebox{1.0\linewidth}{!}{
\begin{tabular}{l c c c c | c c c | c c c}
\hline
\toprule
\multicolumn{2}{c}{}&\multicolumn{3}{c}{A2D2 $\to$ SemanticKITTI}&\multicolumn{3}{c}{Synthia $\to$ SemanticKITTI}&\multicolumn{3}{c}{nuScenes Day $\to$ Night}\\
\cline{3-5}\cline{6-8} \cline{9-11}
Method & Adapt & 2D & 3D & Softmax avg  & 2D & 3D & Softmax avg & 2D & 3D & Softmax avg   \\
\midrule
Source-only & - & 37.4 & 35.3 & 41.5 & 21.1 & 25.9 & 28.2 & 42.2 & 41.2 & 47.8\\
\midrule
xMUDA~\cite{jaritz2020xmuda} & \multirow{2}{*}{UDA} & 36.8 & 43.3 & 42.9 & 25.6 & 30.3 & 33.4 & 46.2 & 44.2 & 50.0 \\
xMUDA$_{PL}$ offline~\cite{jaritz2020xmuda} & & 43.7 & 48.5 & 49.1 & 25.4 & 33.9 & 35.3 & 47.1 & 46.7 & 50.8 \\
\midrule
\midrule
TENT~\cite{wang2021tent} - Eq.\eqref{eqn:ent} & \multirow{7}{*}{TTA} & 39.2 & 36.6 & 40.8 & 25.3 & 23.8 & 27.8 & 39.0 & 43.6 & 43.0 \\
TENT$_{Ens}$ - Eq.\eqref{eqn:ent} & & 39.6 & 36.6 & 41.1 & 27.7 & 23.8 & 29.7 & 39.5 & 43.7 & 43.5\\
xMUDA - Eq.\eqref{eqn:cons} & & 37.5 & 38.0 & 40.2 & 24.0 & 24.1 & 28.0 & 41.7 & 43.9 & 47.0 \\
xMUDA+TENT - Eq.\eqref{eqn:ent},\eqref{eqn:cons} & & 38.1 & 37.5 & 40.5 & 24.4 & 24.0 & 28.0 & 41.8 & \textbf{44.0} & 43.5 \\
xMUDA+TENT$_{Ens}$ - Eq.\eqref{eqn:ent},\eqref{eqn:cons} & & 37.5 & 38.0 & 40.2 & 24.1 & 24.1 & 28.0 & 40.9 & 43.9 & 43.0\\
xMUDA$_{PL}$ - Eq.\eqref{eqn:cons},\eqref{eqn:pseudo} & & 36.5 & 39.5 & 42.9 & 24.2 & 25.0 & 29.0 & 40.8 & 43.6 & 45.2 \\
xMUDA$_{PL}$+TENT$_{Ens}$ - Eq.\eqref{eqn:ent},\eqref{eqn:cons},\eqref{eqn:pseudo} & & 37.0 & 40.0 & 43.0 & 24.3 & 25.0 & 29.0 & 41.3 & 43.6 & 46.0 \\
\midrule
MM-TTA (\textit{Hard Select} - Eq.\eqref{eq:hard})  & \multirow{2}{*}{TTA} & 43.3 & 42.4 & 47.0 & 31.4 & 29.9 & \textbf{35.2} & 42.6 & 43.6 & 51.1\\
MM-TTA (\textit{Soft Select} - Eq.\eqref{eq:soft_1})) & & \textbf{43.7} & \textbf{42.5} & \textbf{47.1} & \textbf{31.5} & \textbf{30.0} & 35.1 & \textbf{44.2} & 43.7 & \textbf{51.8} \\
\bottomrule
\end{tabular}
}
\vspace{-2mm}
\caption{Quantitative comparisons with UDA methods and TTA baselines for multi-modal 3D semantic segmentation.}
\label{table:main_table}
\vspace{-3mm}
\end{table*}

\subsection{Datasets and Settings}
We evaluate our proposed MM-TTA on several scenarios where test-time adaptation is necessary. 
First, sensor setups of camera and LiDAR are different between training and test data in real-world, where we adopt the benchmark \textit{\textbf{A2D2-to-SemanticKITTI}}. In particular, A2D2~\cite{geyer2020a2d2} provides a 2.3 MegaPixels (MP) camera and 16 channels of LiDAR, while SemanticKITTI~\cite{behley2019semantickitti} uses a 0.7MP camera and 64 channels of LiDAR. This difference in hardware specification can cause unpredictable domain shift in the real-world so that the pre-trained model on source needs to be quickly adapted to the incoming test data.
Second, another real-world case is \textit{\textbf{nuScenes Day-to-Night}}, where we use nuScenes~\cite{caesar2020nuscenes} for this adaptation scenario. LiDAR is an active sensor that emits laser beams that are mostly invariant to lighting conditions.
However, images captured by day and night are obviously different in color distribution, leading to a performance degradation without any adaptation.

Finally, we evaluate test-time adaptation between synthetic and real data using \textit{\textbf{Synthia-to-SemanticKITTI}}, which is a challenging benchmark that needs to handle a significant domain shift not only in camera (style gap due to the lack of photorealism in synthetic data) but also in LiDAR (point distribution and depth accuracy). 

For A2D2-to-SemanticKITTI and nuScenes Day-to-Night, we follow the dataset setting in xMUDA~\cite{jaritz2020xmuda}.
For Synthia-to-SemanticKITTI, we newly organize Synthia~\cite{Ros_2016_CVPR} by constructing point clouds with provided image and depth ground truth. Since depth maps are dense, we randomly sample pixels to obtain corresponding point clouds. Details are provided in the supplementary material.

\subsection{Implementation Details}
\noindent\textbf{Multi-modal model:} we follow xMUDA~\cite{jaritz2020xmuda} to construct the two-stream multi-modal framework. For the 2D branch, we adopt U-Net~\cite{ronneberger2015unet} with a ResNet34~\cite{he2015deep} encoder.
For the 3D branch, we use a U-Net (downsampling 6-times) that utilizes sparse convolution~\cite{SubmanifoldSparseConvNet} on the voxelized point cloud input, where we use either SparseConvNet~\cite{graham20173d} or MinkowskiNet~\cite{choy20194d} for our settings\footnote{For the A2D2-to-SemanticKITTI and Synthia-to-SemanticKITTI settings, we find that 
there is a reported implementation issue of SparseConvnet in xMUDA,
and thus we use MinkowskiNet in other 3D segmentation repostitory~\cite{tang2020searching} for stability. For nuScenes Day-to-Night, we use SparseConvNet as in xMUDA~\cite{jaritz2020xmuda}.}. 
%
For each setting, all the baseline comparisons are evaluated using the same framework and backbone models.


\vspace{2mm}
\noindent\textbf{Pre-training with source data:} we directly utilize the source pre-trained model from the xMUDA official code for fair comparisons when we use the SparseConvnet. On the other hand, we train the MinkowskiNet on source data from scratch. To reproduce similar performance only using the source as in xMUDA, we use the Adam optimizer with learning rate of 1x10$^{-3}$ for the 2D model, and SGD momentum with learning rate of 2.4x10$^{-1}$ for the 3D model.

\vspace{2mm}
\noindent\textbf{Test-time adaptation on target data:} TTA~\cite{wang2021tent} only optimizes for batch norm affine parameters during training and then reports performance after 1 epoch of adaptation. We adopt the same setting for all the baselines and our method, where we use batch statistics to compute the normalization parameters at test time. To implement our slow-fast modeling strategy in Intra-PG, we first copy the source pre-trained model and then gradually update batch norm statistics during adaptation with a momentum from the fast model.

\begin{figure*}[!t]
\vspace{-2mm}
\begin{center}
\includegraphics[width=0.9\linewidth]{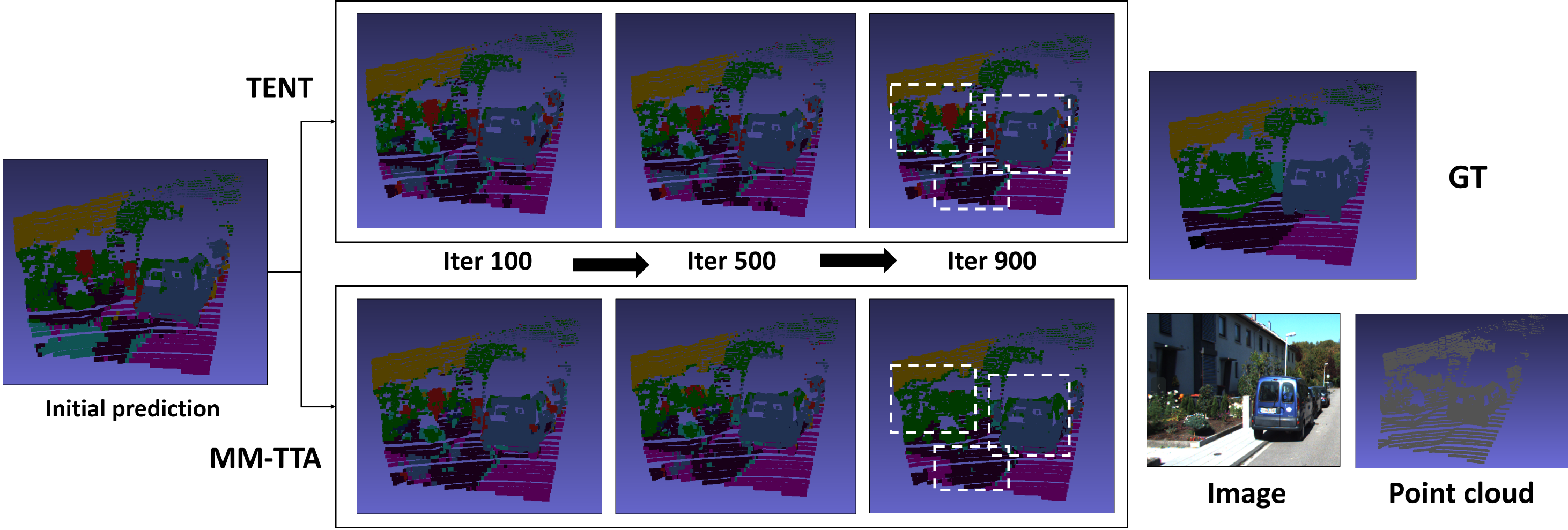}
\caption{\textbf{Example results of our MM-TTA during test-time adaptation for gradual improvement.} 
While TENT~\cite{wang2021tent} shows little improvements during adaptation, our method can effectively suppress the noise and achieve visually similar results to the ground truth, especially within the area of dotted white boxes.
}
\label{fig:visual_comparison}
\end{center}
\vspace{-5mm}
\end{figure*}

\begin{table*}[!t]
\centering
    \resizebox{0.94\linewidth}{!}{
\begin{NiceTabular}{l c | c c | c c | c | c c c}
\hline
\toprule
& \multirow{2}{*}{Method} & \multicolumn{2}{c}{Intra-PG} & \multicolumn{2}{c}{Inter-PR} & \multirow{2}{*}{Thres. on pseudo-label} & \multirow{2}{*}{2D} & \multirow{2}{*}{3D} & \multirow{2}{*}{Softmax avg} \\
\cline{3-4}\cline{5-6}
& & Fast & Slow & Fusion & Select & & & & \\
\midrule
\multirow{9}{*}{Pseudo} & (1) & \checkmark & & & & & 39.2 & 36.7 & 40.8 \\
& (2) & \checkmark & \checkmark & & & & 40.1 & 37.6 & 41.9 \\
& (3) & \checkmark& \checkmark & Consensus & & & 40.8 & 39 & 41.8 \\
& (4) & \checkmark& \checkmark & Consensus & & \checkmark & 40.8 & 37.5 & 43.8 \\
& (5) & \checkmark & \checkmark & Merge & & \checkmark & 43.1 & 37.4 & 45.3 \\
& (6) & \checkmark &  & Merge & & \checkmark & 39.3 & 36.7 & 41.6 \\
& (7) & \checkmark & \checkmark & & Entropy & \checkmark & 40.2 & 39.6 & 43.4\\ 
& MM-TTA & \checkmark & \checkmark & & Consistency (Hard) & \checkmark & 43.3 & 42.4 & 47.0\\ 
& MM-TTA & \checkmark & \checkmark & & Consistency (Soft) & \checkmark & 43.7 & 42.5 & 47.1 \\
\bottomrule
\end{NiceTabular}
}
\vspace{-1mm}
\caption{
\textbf{Ablation study on effects of Intra-PG and Inter-PR in the A2D2 $\rightarrow$ SemanticKITTI benchmark.}
We provide two variants with different fusion: 1) Consensus: using pseudo-labels that are consistent between 2D and 3D, and 2) Merge: taking the mean of two output probabilities. For the selection process, ``Entropy'' calculates and  compares  the  entropy  of  2D  and  3D  predictions.
}
\label{table:ablation}
\vspace{-3mm}
\end{table*}

\subsection{Main Results}
In this section, we show quantitative evaluations on the aforementioned three benchmark settings by reporting mIoU on predictions of 2D, 3D and ensembling between their predicted probabilities (see \Tref{table:main_table}).
For each benchmark setting, we mainly compare our method with Test-Time Adaptation (TTA) baselines, while also reporting results for xMUDA that uses Unsupervised Domain Adaptation (UDA) as references, which can access both source and target data without the budget constraint during training.

\vspace{1mm}
\noindent\textbf{Baselines.}
For UDA, we compare with the multi-modal xMUDA framework that utilizes consistency loss (\textbf{xMUDA}) and self-training using offline pseudo-labels (\textbf{xMUDA$_{PL}$ offline}).
For TTA baselines, we evaluate \textbf{TENT}, \textbf{xMUDA}, \textbf{xMUDA$_{PL}$}, as introduced in~\cref{sec:baseline}. 
%
Then, we extend TENT to multiple modalities (\textbf{TENT$_{Ens}$}), where we do entropy minimization on the ensemble of 2D and 3D logits.
We also include the combinations of these methods, xMUDA+TENT, xMUDA+TENT$_{Ens}$, and xMUDA$_{PL}$+TENT$_{Ens}$.
For all methods, we do a hyperparameter search and report best results.

\vspace{1mm}
\noindent\textbf{Results.}
In \Tref{table:main_table}, we show that our MM-TTA methods (both \textit{Hard Select} and \textit{Soft Select}) perform favorably against all the TTA baselines in three benchmark settings. For TTA baselines on A2D2-to-SemanticKITTI and Synthia-to-SemanticKITTI, we find that entropy and pseudo-labeling based methods (e.g., TENT, xMUDA$_{PL}$) perform better than the consistency loss (e.g., xMUDA), due to the difficulty of capturing the correct consistency across modalities.
In addition, although some TTA baselines (\eg, TENT$_{Ens}$, xMUDA$_{PL}$) improve the performance of individual 2D and 3D predictions, the ensemble results are all worse than the ``Source-only'' model. 
This is because these methods do not have a well-designed module to jointly consider multi-modal outputs, where we use our Inter-PR to adaptively generate cross-modal pseudo-labels.
%

For nuScenes Day-to-Night, different from the other settings, the domain gap is larger for RGB than for LiDAR, and thereby the challenge mainly lies in how to improve the 2D branch and obtain effective ensemble results.
For all the baselines and our methods, IoU for the 3D branch is competitive, while our results in the 2D branch and the ensemble are significantly improved, which shows the benefits of our designed Intra-PG and Inter-PR modules. Surprisingly, the ensemble results of our MM-TTA methods are better than the ones in the xMUDA approaches that use the UDA setting. This shows the effectiveness of our proposed MM-TTA framework for fast test-time adaptation.
\Fref{fig:visual_comparison} show example results of 3D semantic segmentation on SemanticKITTI. Our MM-TTA method gradually improves the initial prediction throughout adaptation, and produces more complete and accurate outputs compared to TENT. 





\begin{table}[!t]
\vspace{-2mm}
\centering
\vfill
\begin{subtable}{0.43\textwidth}{\renewcommand\arraystretch{1.0}
    \tiny
    \centering
        \vspace{2mm}
\resizebox{\textwidth}{!}{
    \begin{tabular}{l c c c c}
    \hline
    \toprule
    Method  & Threshold & 2D & 3D & Softmax avg \\
    \midrule
    \multirow{4}{*}{Hard} & 0.1 & 40.3 & 41.3 & 45.2\\
    & 0.3 & 43.3 & 42.4 & 47.0\\
    & 0.5 & 43.2 & 42.5 & 46.7\\
    & 0.7 & 43 & 42.3 & 46.2\\
    \midrule
    \multirow{4}{*}{Soft} & 0.1 & 41.2 & 41.5 & 45.7\\
    & 0.3 & 43.7 & 42.5 & 47.1\\
    & 0.5 & 43.9 & 42.6 & 46.9\\
    & 0.7 & 43.7 & 42.3 & 46.3\\
    \bottomrule
    \end{tabular}
    }
    }
   \caption{Pseudo-label threshold ratio $\theta^{(k)}$}
    \label{tab:pseudo_cut}
\end{subtable}%
\vfill
\begin{subtable}{0.4\textwidth}{\renewcommand\arraystretch{1.0}
    \small
    \centering
        \vspace{2mm}
\resizebox{\textwidth}{!}{
    \begin{tabular}{l c c c c}
    \hline
    \toprule
    Method  & Momentum & 2D & 3D & Softmax avg \\
    \midrule
    \multirow{3}{*}{Hard} & 1.00 & 42.8 & 42.0 & 46.3\\
    & 0.99 & 43.3 & 42.4 & 47.0\\
    & 0.95 & 42.0 & 42.2 & 46.1\\
    \midrule
    \multirow{3}{*}{Soft} & 1.00 & 43.2 & 42.1 & 46.5\\
    & 0.99 & 43.7 & 42.5 & 47.1\\
    & 0.95 & 42.6 & 42.4 & 46.3\\
    \bottomrule
    \end{tabular}
      }
      }
        \caption{Momentum factor $\lambda$}
    \label{tab:momen_up}
\end{subtable}%
\vspace{-3mm}
\caption{Sensitivity analysis in A2D2 $\rightarrow$ SemanticKITTI.
}
\label{tab:abla}
\vspace{-5mm}
\end{table}

\subsection{Ablation Study}
\subsubsection{Analysis on MM-TTA}


\vspace{1mm}
\noindent\textbf{Inter-PR for pseudo-label refinement.}
We show different pseudo-label refinement methods for Inter-PR, and compare them with our \textit{Hard Select} and \textit{Soft Select} schemes. First, in Method (4) and (5) of \Tref{table:ablation} respectively, we use two simple fusion techniques: 1) only using points that are consistent between pseudo-labels of 2D and 3D (Consensus), and 2) taking the mean of two output probabilities for pseudo-labeling (Merge). Second, for selecting pseudo-labels from either the 2D or 3D branch, one alternative is to calculate and compare the entropy of 2D and 3D predictions (Entropy) as in Method (7).
Overall, our MM-TTA methods perform better than these model variants. In addition, we show that using the threshold on pseduo-labels is a good choice, e.g., comparing Method (3) with (4).


\vspace{1mm}
\noindent\textbf{Intra-PG with slow-fast modeling.}
We design model variants to validate the effectiveness of Intra-PG. In Method (2)/(5) of \Tref{table:ablation}, using the slowly-updated model improves Method (1)/(6), respectively. This shows that Intra-PG is useful with different pseudo-labeling schemes, e.g., without fusion in Method (2) or ``Merge'' in Method (5). Note that our Inter-PR module requires slow-fast modeling and thus these two modules are coupled together as our final model, which shows performance gains compared to other variants.

\vspace{-2mm}
\subsubsection{Sensitivity Analysis}
\noindent\textbf{Threshold $\theta^{(k)}$.} 
This threshold is critical for pseudo-labeling, where low values filter more points in a class-wise manner, and vice versa. \Tref{tab:pseudo_cut} shows the robustness of our method to $\theta^{(k)}$, and a value of 0.3 performs best.

\vspace{1mm}
\noindent\textbf{Momentum factor $\lambda$.}
We use a slow-fast modeling strategy to slowly update the source pre-trained batch norm statistics with a momentum $\lambda$ from the fast-updated model. \Tref{tab:momen_up} shows the effect of changing $\lambda$. Setting it as 1.0 would simply keep the source statistics and is not optimal.

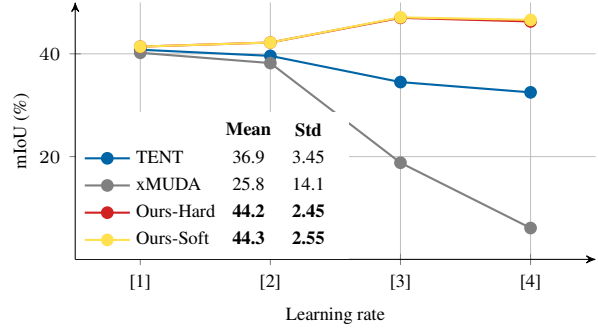
\begin{figure}[!t]
\begin{center}
\pgfplotsset{
    legend image with text/.style={
        legend image code/.code={%
            \node[anchor=center] at (0.3cm,0cm) {#1};
        }
    },
}
\begin{tikzpicture}[>=stealth']
\begin{axis}[
    width=8.5cm,
    height=5cm,
    grid=major,
    major grid style={line width=.2pt,draw=gray!50},
    xmin=0.5,
    xmax=4.5,
    ymin=0,
    ymax=50,
    axis lines=middle,
    axis line style={->},
    x label style={at={(axis description cs:0.5,-0.15)},anchor=north},
    y label style={at={(axis description cs:-0.07,.5)},rotate=90,anchor=south},
    y tick label style={
        /pgf/number format/.cd,
        fixed,
        fixed zerofill,
        precision=0,
        /tikz/.cd
    },
    xlabel={Learning rate},
    xtick={1,2,3,4},
    xticklabels={[1],[2],[3],[4]},
    ylabel={mIoU (\%)},
    legend style={at={(0.0, 0.0)}, anchor=south west, draw=none},
    legend cell align={left},
    legend columns=3,
]
\addlegendimage{legend image with text=};
\addlegendentry{};
\addlegendimage{legend image with text=\textbf{Mean}};
\addlegendentry{};
\addlegendimage{legend image with text=\textbf{Std}};
\addlegendentry{};
\addplot[color=plotsblue,solid,thick,mark=*, mark options={solid}] coordinates {
    (1, 40.8)
    (2, 39.6)
    (3, 34.5)
    (4, 32.5)
};
\addlegendentry{TENT};
\addlegendimage{legend image with text=36.9};
\addlegendentry{};
\addlegendimage{legend image with text=3.45};
\addlegendentry{};
\addplot[color=plotsgrey,solid,thick,mark=*, mark options={solid}] coordinates {
    (1, 40.2)
    (2, 38.2)
    (3, 18.8)
    (4, 6.1)
};
\addlegendentry{xMUDA};
\addlegendimage{legend image with text=25.8};
\addlegendentry{};
\addlegendimage{legend image with text=14.1};
\addlegendentry{};
\addplot[color=plotsred,solid,thick,mark=*, mark options={solid}] coordinates {
    (1, 41.4)
    (2, 42.2)
    (3, 47.0)
    (4, 46.3)
};
\addlegendentry{Ours-Hard};
\addlegendimage{legend image with text=\textbf{44.2}};
\addlegendentry{};
\addlegendimage{legend image with text=\textbf{2.45}};
\addlegendentry{};
\addplot[color=plotsyellow,solid,thick,mark=*, mark options={solid}] coordinates {
    (1, 41.4)
    (2, 42.2)
    (3, 47.1)
    (4, 46.6)
};
\addlegendentry{Ours-Soft}
\addlegendimage{legend image with text=\textbf{44.3}};
\addlegendentry{};
\addlegendimage{legend image with text=\textbf{2.55}};
\addlegendentry{};
\end{axis}
\end{tikzpicture}
\vspace{-3mm}
\caption{
\textbf{Stability on using different learning rates in A2D2 $\rightarrow$ SemanticKITTI.} For the 2D/3D branch, we use four sets of learning rates: [1] 1.0x10$^{-5}$/2.4x10$^{-5}$, [2] 1.0x10$^{-5}$/2.4x10$^{-4}$, [3] 1.0x10$^{-4}$/2.4x10$^{-4}$, [4] 1.0x10$^{-4}$/2.4x10$^{-3}$.
}
\label{fig:lr_graph}
\end{center}
\vspace{-5mm}
\end{figure}

\vspace{1mm}
\noindent\textbf{Stability during TTA.}
Since TTA only sees the test data once during adaptation, the stability can be largely affected by hyperparameters like the learning rate. In \Fref{fig:lr_graph}, we run different methods with various learning rates, and find that our MM-TTA methods perform robustly and show a good stability during adaptation with the result of higher mean (44.2/44.3) and lower standard deviation (2.45/2.55). 

\begin{figure}[!t]
\begin{center}
\begin{tikzpicture}[>=stealth']
\begin{axis}[
    width=8.5cm,
    height=4cm,
    grid=major,
    major grid style={line width=.2pt,draw=gray!50},
    xmin=50,
    xmax=650,
    ymin=62,
    ymax=76,
    axis lines=middle,
    axis line style={->},
    x label style={at={(axis description cs:0.98,0.03)},anchor=south east},
    y label style={at={(axis description cs:-0.07,.5)},rotate=90,anchor=south},
    y tick label style={
        /pgf/number format/.cd,
        fixed,
        fixed zerofill,
        precision=0,
        /tikz/.cd
    },
    xlabel={Iteration number},
    ylabel={Accuracy (\%)},
    legend style={at={(0.0, -0.2)}, anchor=north west, draw=none},
    legend cell align={left},
    legend columns=2,
]
\addplot[color=plotsblue,solid,thick,mark=*, mark options={solid}] coordinates {
    (100, 63.85)
    (200, 64.55)
    (300, 64.8)
    (400, 66.3)
    (500, 66.25)
    (600, 66.7)
};
\addlegendentry{Fast};
\addplot[color=plotsorange,solid,thick,mark=*, mark options={solid}] coordinates {
    (100, 66.15)
    (200, 66.75)
    (300, 67.85)
    (400, 69.75)
    (500, 69.35)
    (600, 69.9)
};
\addlegendentry{Fast+Slow (Intra-PG)};
\addplot[color=plotsgrey,solid,thick,mark=*, mark options={solid}] coordinates {
    (100, 71.3)
    (200, 73.1)
    (300, 73.8)
    (400, 75.1)
    (500, 74.9)
    (600, 75.1)
};
\addlegendentry{Intra-PG+Inter-PG (Hard)}
\addplot[color=plotsyellow,solid,thick,mark=*, mark options={solid}] coordinates {
    (100, 71.4)
    (200, 73.3)
    (300, 74.1)
    (400, 75.3)
    (500, 75.0)
    (600, 75.1)
};
\addlegendentry{Intra-PG+Inter-PG (Soft)}
\end{axis}
\end{tikzpicture}
\vspace{-4mm}
\caption{
Pseudo-label accuracy during adaptation in A2D2 $\rightarrow$ SemanticKITTI.
}
\label{fig:pseudo_acc}
\end{center}
\vspace{-5mm}
\end{figure}
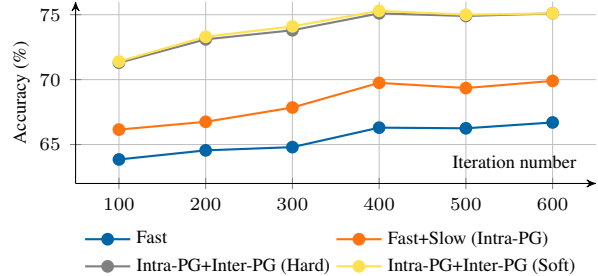

\vspace{-2mm}
\subsubsection{Analysis on Pseudo-labeling Accuracy}
We measure the pseudo-label accuracy at different iterations during adaptation for our proposed modules. We test 6 phases from the iterations of 100 to 600.
In each phase, we collect pseudo-labels for valid points and calculate the average accuracy over all categories.
In \Fref{fig:pseudo_acc}, we first observe that using slow-fast modeling in Intra-PG improves the accuracy from the baseline (only using the fast model) by 2\%. Then, combining our proposed two modules consistently shows improvement in all iterations, with a 5\% gain.

\vspace{-1mm}
\section{Conclusions}
In this paper, we present a new problem setting, Multi-Modal Test-Time Adaptation (MM-TTA) for 3D semantic segmentation.
We first identify several baselines and their limitations, and then propose a simple yet effective self-training framework consisting of two modules, Intra-PG and Inter-PR, to produce reliable cross-modal pseudo-labels.
In experiments, we demonstrate our MM-TTA framework in several benchmark settings. In addition, we provide extensive ablation studies and analysis to show the benefits of our proposed modules.

{
\noindent\textbf{Acknowledgement}
This work was part of Inkyu Shin’s internship at NEC Laboratories America and was also partially supported by Samsung Electronics Co., Ltd (G01200447), and under the framework of international cooperation program managed by the National Research Foundation of Korea (NRF-2020M3H8A1115028, FY2021).
}

{
\small
\bibliographystyle{cvpr}
\bibliography{cvpr}
}

\clearpage

\appendix
\section*{Appendix}

\noindent In this appendix, we provide,

\begin{enumerate}[label=\Alph*)]
    \item Dataset construction
    \item Algorithm for our MM-TTA
    \item Oracle test
    \item Analysis on class-wise adaptation
    \item Qualitative results for pseudo labels
    \item More qualitative results
    \item Limitations
\end{enumerate}

\section{Dataset Construction}

\vspace{-0.5mm}
\subsection{A2D2, SemanticKITTI and nuScenes}
We strictly follow the dataset setting of A2D2/SemanticKITTI/nuScenes from xMUDA~\cite{jaritz2020xmuda}.
For A2D2 and SemanticKITTI, 10 classes are shared and used to train and test in each dataset. 
The 10 classes are car, truck, bike, person, road, parking, sidewalk, building nature, other-objects.

\vspace{-0.5mm}
\subsection{Synthia}
\begin{figure}[t]
\begin{center}
\includegraphics[width=1.0\linewidth]{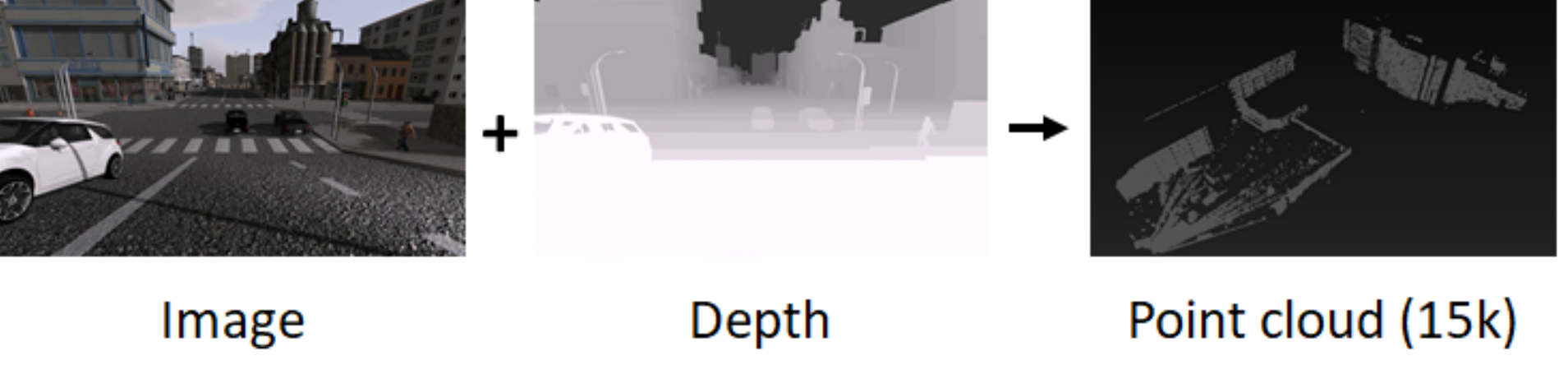}
\caption{Construction of the Synthia dataset to generate point clouds (15k points). In that sense, we can simulate the multi-modal dataset.}
\label{fig:synthia}
\end{center}
\vspace{-5mm}
\end{figure}

We re-organize the synthia dataset~\cite{Ros_2016_CVPR} to simulate synthetic-to-real setting. It initially contains 9,000 RGB images with corresponding labels of segmentation and depth. Since every pixel in RGB images can be formatted into point cloud with depth ground truth, we randomly sample about 15k number of pixels to make up the point cloud of that scene (see ~\Fref{fig:synthia}). We use all of the data as the training set and merge 23 classes into 10 categories to be shared with the SemanticKITTI dataset. 
%

\section{Algorithm for MM-TTA}
Here, we provide an algorithm for MM-TTA consisting of the proposed two modules: Intra-PG and Inter-PR in~\Aref{alg:pseudo}.

\section{Oracle Test}
We provide two kinds of oracle tests: 1) Oracle TTA: only BN parameters of 2D/3D models are finetuned using the real target label during 1 epoch. 2) Oracle Full: all layers are updated with real target label from scratch during 30 epochs. In both cases, our MM-TTA is able to generate reliable pseudo labels, where the performance is achieved closer to that of oracle. Specifically, on nuScenes Day $\to$ Night, our MM-TTA using \textit{Soft Select} obtains comparable results to ``Oracle TTA'' that uses real labels.

\begin{table*}
\centering
\resizebox{1.0\linewidth}{!}{
\begin{tabular}{l c c c c | c c c | c c c}
\hline
\toprule
\multicolumn{2}{c}{}&\multicolumn{3}{c}{A2D2 $\to$ SemanticKITTI}&\multicolumn{3}{c}{Synthia $\to$ SemanticKITTI}&\multicolumn{3}{c}{nuScenes Day $\to$ Night}\\
\cline{3-5}\cline{6-8} \cline{9-11}
Method & Adapt & 2D & 3D & Softmax avg  & 2D & 3D & Softmax avg & 2D & 3D & Softmax avg   \\
\midrule
Source-only & - & 37.4 & 35.3 & 41.5 & 21.1 & 25.9 & 29.3 & 42.2 & 41.2 & 47.8\\
\midrule

TENT~\cite{wang2021tent} & \multirow{3}{*}{TTA} & 39.2 & 36.6 & 40.8 & 25.3 & 23.8 & 27.8 & 39.0 & 43.6 & 43.0 \\
MM-TTA (\textit{Hard Select})  & & 43.3 & 42.4 & 47.0 & 31.4 & 29.9 & 35.2 & 42.6 & 43.6 & 51.1\\
MM-TTA (\textit{Soft Select}) & & 43.7 & 42.5 & 47.1 & 31.5 & 30.0 & 35.1 & 44.2 & 43.7 & 51.8 \\

\midrule
\midrule

Oracle TTA & TTA & 48.5 & 45.8 & 52.4 & 38.8 &  31.1 & 41.4 & 45.6 & 43.6 & 51.5 \\

Oracle Full & - & 57.9 & 66.6 & 69.5 & 57.9 & 66.6 & 69.5 & 48.6 & 47.1 & 55.2 \\
\bottomrule
\end{tabular}
}
\vspace{-2mm}
\caption{Quantitative results with using real target labels as oracles. Depending on whether we only finetune the batchnorm parameters or update all layers, the oracles are ``Oracle TTA'' and ``Oracle Full''.}
\label{table:orcle_test}
\end{table*}

\begin{algorithm*}[hbt!]
\DontPrintSemicolon
\vspace{-0.5mm}
\caption{Algorithm for MM-TTA}
\label{alg:pseudo}
\KwIn{Target data $x_{t}$ = ($x^{2D}_{t}$, $x^{3D}_{t}$), Source pre-trained model $F^M$ = ($F^{2D}$, $F^{3D}$)}
\KwOut{The model with adapted weights on the target dataset $F^{M}$ = ($F^{2D}$, $F^
{3D}$)}
\Begin
{
    Define the slow model and copy weights \\
    $F \leftarrow S$ \\
    $F$.train(), $S$.eval() \\
    \For{1 epoch}
        {
            \textcolor{teal}{\# 1.Intra-PG}\\
            Fuse the outputs of slow-fast model (Eq.(\textcolor{red}{7})). \\
            $p(x^{M}_{t}) = Fuse(S^{M}(x^{M}_{t}),F^{M}(x^{M}_{t})).$ \\
            Obtain aggregated pseudo labels (Eq.(\textcolor{red}{8})).\\
            $\hat{y}^{M}_{t} = \argmax_{k \in K} p(x^{M}_{t})^{(k)}$ \\ 
            \textcolor{teal}{\# 2.Inter-PR}\\
            Calculate a consistency measure between slow and fast models (Eq.(\textcolor{red}{9}), (\textcolor{red}{10})).\\
            $\zeta_{M} = Sim(S^{M}(x^{M}_{t}), F^{M}(x^{M}_{t}))$ \\
            \If{Hard Select}
            {
                Select from one of the modalities (Eq.(\textcolor{red}{11}).\\
                $\hat{y}_{t}^{Ens} = \begin{cases}
                \hat{y}^{\textrm{2D}}_{t}, & \text{if $\zeta_{\textrm{2D}} \geq \zeta_{\textrm{3D}}$}, \\
                \hat{y}^{\textrm{3D}}_{t}, & \text{otherwise}. \\ 
                \end{cases}$
            }
            \ElseIf{Soft Select}
            {
                Weighted sum from the two modalities (Eq.(\textcolor{red}{12}), (\textcolor{red}{13})). \\
                $\hat{y}_{t}^{Ens} = \argmax_{k \in K}p^{W(k)}_{t}$ \\
                $p^{W(k)}_{t}$ = $Weight(p(x^{\textrm{2D}}_{t})^{(k)}, p(x^{\textrm{3D}}_{t})^{(k)})$
            }
            \textcolor{teal}{\# 3.Update the model}\\
            Update the $\Omega^{F}$ with Eq.(\textcolor{red}{14}). \\
            Momentum update for $\Omega^{S}$ with Eq.(\textcolor{red}{6})\\
            $\Omega^{S}_{t_{i}} = (1-\lambda)\Omega^{F}_{t_{i}} + \lambda\Omega^{S}_{t_{i-1}}$
        }
        
}
\end{algorithm*}

\section{Analysis on Class-wise Adaptation}
\begin{figure*}[t]
\begin{center}
\includegraphics[width=1.0\linewidth]{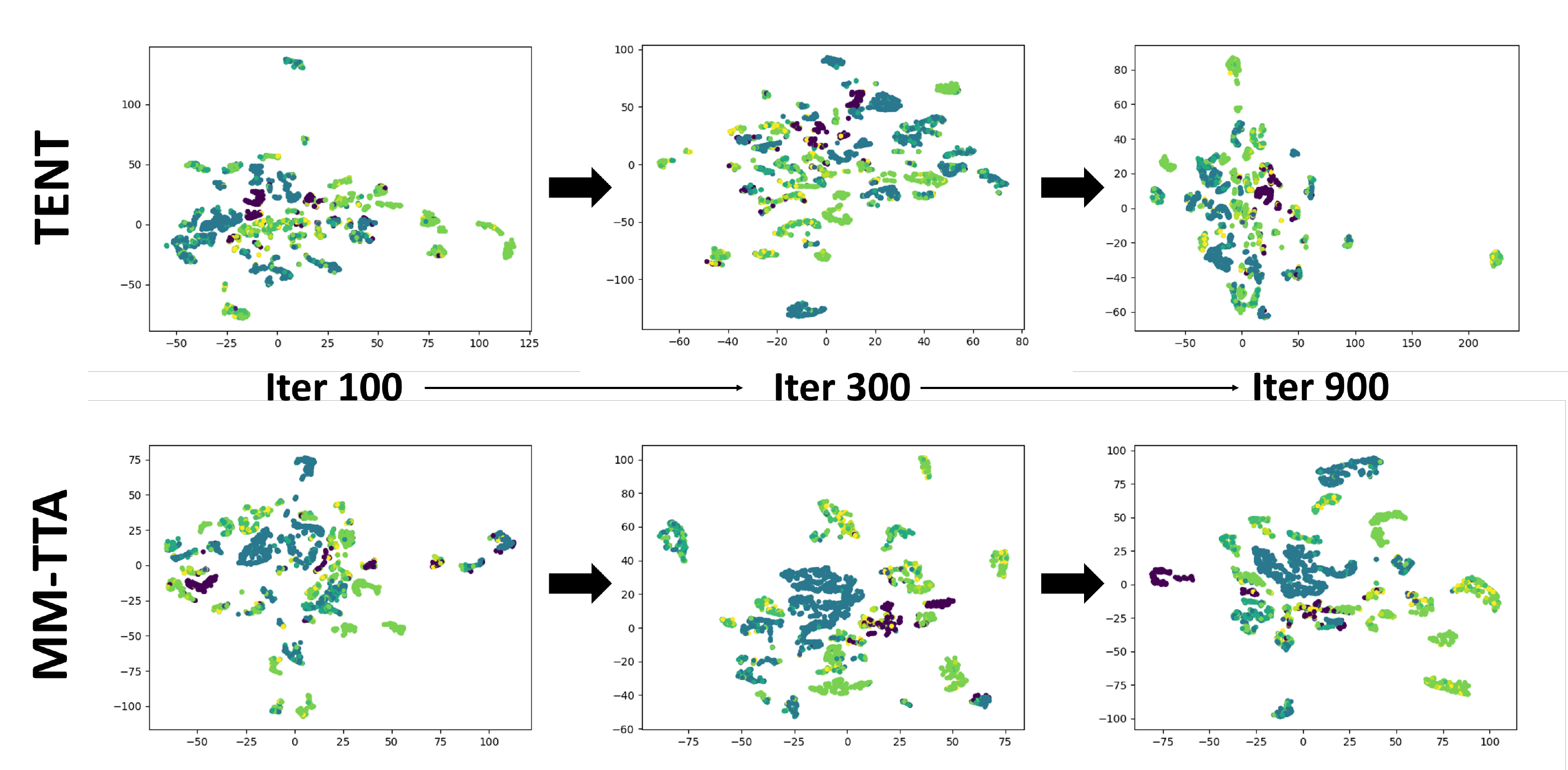}
\caption{Qualitative results of t-SNE on TENT and MM-TTA. Each color represents one category, where our MM-TTA produces more compact clusters for each category}
\label{fig:tsne}
\end{center}
\vspace{-5mm}
\end{figure*}

We analyze how the test-time adaptation has class-wise effect while proceeding with the iterations. We visualize class-wise adaptation with t-SNE~\cite{tsne} and conduct an analysis comparing between our MM-TTA and the TENT baseline (see~\Fref{fig:tsne}). We map the final logit of all of test data's points to the 2-D space via t-SNE. We observe that our MM-TTA performs better category-level feature alignment during test-time adaptation at across different iterations. 

\section{Qualitative Results for Pseudo Labels}
\begin{figure*}[t]
\begin{center}
\includegraphics[width=1.05\linewidth]{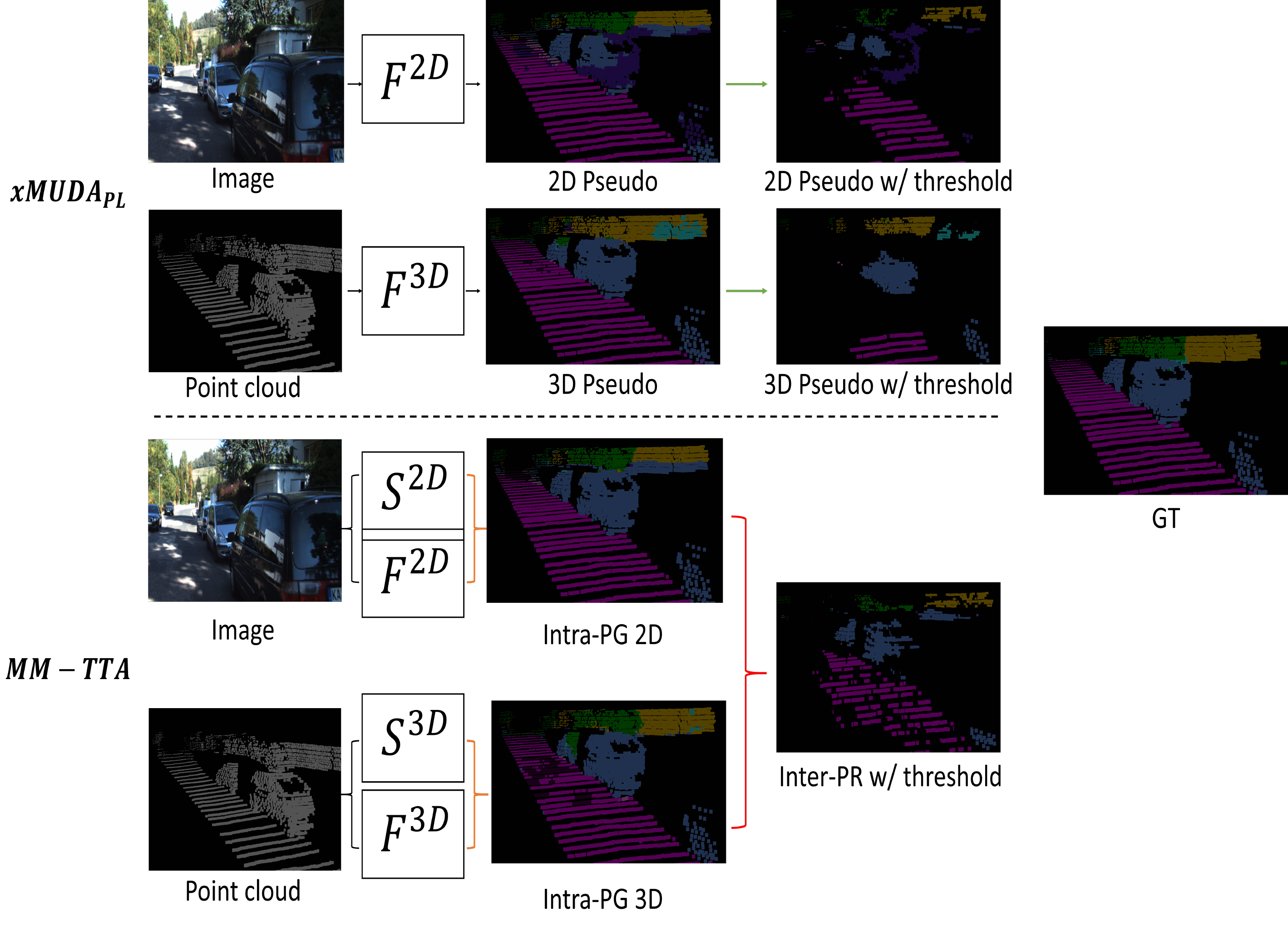}
\caption{Qualitative results of pseudo labels on xMUDA$_{PL}$ and MM-TTA.}
\label{fig:pl_vis}
\end{center}
\end{figure*}

In~\Fref{fig:pl_vis}, we visualize the pseudo labels generated from MM-TTA and compare with other baselines. We can find that our two modules achieve more refined and accurate pseudo label that is more similar to GT.

\section{More Qualitative Results}
Given several multi-modal data with image and point cloud, we visualize the qualitative 3D segmentation results of our MM-TTA and compare with other baselines (TENT and xMUDA). In all adaptation scenarios, we can observe that our MM-TTA (both \textit{Hard Select} and \textit{Soft Select}) shows more similar results to ground truth (GT).

\begin{figure*}[t]
\begin{center}
\includegraphics[width=1.05\linewidth]{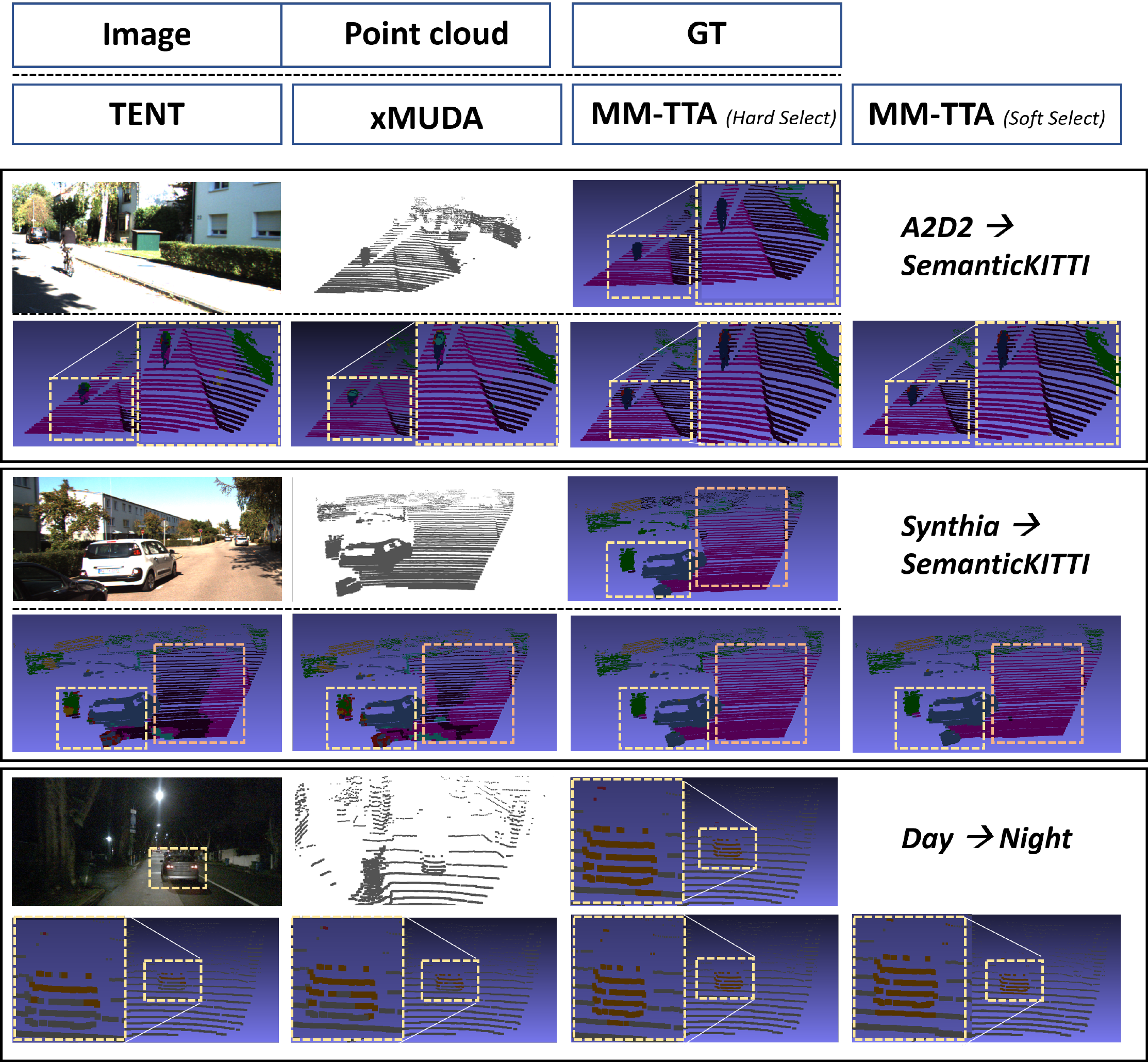}
\caption{Qualitative 3D segmentation result of TENT, xMUDA, MM-TTA (\textit{Hard Select}), MM-TTA (\textit{Soft Select}) on three adaptation benchmarks.}
\label{fig:qual_add}
\end{center}
\end{figure*}

\section{Limitations and Discussion}
Since our method focuses on selecting or giving adaptive weights between two modalities for general pseudo-label generation, one limitation is that its effectiveness may vary across categories. Therefore, one future direction is to develop category-aware test-time adaptation methods, so that the model can further boost the performance for certain classes that do not perform well.


\end{document}